\documentclass[12pt]{article}

\usepackage[english]{babel}
\usepackage{authblk}
\usepackage[a4paper,top=2.5cm,bottom=2.5cm,left=3cm,right=3cm,marginparwidth=2.75cm]{geometry}
\usepackage{marginnote}
\usepackage{amsmath,amssymb}
\usepackage[english]{babel}
\usepackage{amsthm}
\usepackage{natbib}
\usepackage{graphicx}
\usepackage[colorlinks=true, allcolors=blue]{hyperref}
\usepackage{algorithm}
\usepackage{algpseudocode}
\usepackage{wrapfig}

\title{Reward is not enough: can we liberate AI from the reinforcement learning paradigm?}

\author{Vacslav Glukhov}
\affil{glukhov@stanfordalumni.org}

\begin{document}
\maketitle

\begin{abstract}

I present\footnote{This is the 2024 version of the paper originally published in 2022 that fixes most typos and slightly improves the readability of the text. The original version was put together in haste and I apologize to the few readers for my bugs and sloppy phrasing.} arguments against the hypothesis put forward by Silver, Singh, Precup, and Sutton \cite{SILVER2021103535}: reward maximization is not enough to explain many activities associated with natural and artificial intelligence including knowledge, learning, perception, social intelligence, evolution, language, generalisation and imitation. 

I show such \emph{reductio ad lucrum} has its intellectual origins in the political economy of \emph{Homo economicus} and substantially overlaps with the radical version of behaviourism. 

I show why the reinforcement learning paradigm, despite its demonstrable usefulness in some practical applications, is an incomplete framework for intelligence -- natural and artificial.

Complexities of intelligent behaviour are not simply second-order \emph{complications} on top of reward maximisation. This fact has profound implications for the development of practically usable, smart, safe and robust artificially intelligent agents. 
\end{abstract}


\section{Introduction}
\label{sec:intro}
The recent paper from DeepMind \cite{SILVER2021103535} is a disappointment for a reason: the organization is known for their high quality research and "Reward is enough" not their deepest. 

It seems wrong, however, to dismiss it as yet another example of the law of the instrument\footnote{"I suppose it is tempting, if the only tool you have is a hammer, to treat everything as if it were a nail" (attributed to Abraham Maslow)}. Instead, it makes sense to use the opportunity for a broad discussion of intelligence - artificial and natural -- and the principle of reward maximisation. 

A charitable approach to "Reward is enough" is that it might have been written as deliberately provocative -- perhaps to trigger a broader discussion about the principle of reward maximisation and its role in the study of intelligence, natural and artificial. I take up the challenge.

\section{The kitchen robot example}
To put our discussion on a productive path let us use one of the authors' examples. 

 \cite{SILVER2021103535} hypothesize that a singular reward of "cleanliness" must lead a robot to develop advanced perception capabilities and produce complex intelligent behaviours in a particular environment:
\begin{quote}
    In order for a kitchen robot to maximise cleanliness, it must presumably have abilities of perception (to differentiate clean and dirty utensils), knowledge (to understand utensils), motor control (to manipulate utensils), memory (to recall locations of utensils), language (to predict future mess from dialogue), and social intelligence (to encourage young children to make less mess). A behaviour that maximises cleanliness must therefore yield all these abilities in service of that singular goal. 
\end{quote}

No, it must not. 

Maximising "cleanliness" could also yield a robot that destroys all the sources of mess -- utensils, pets, children, along their parents -- minimizing its own cost and maximizing cleanliness, once and forever. 

In order to avoid such disastrous result the robot's designer must impose a set of rules specific to its objective which constrain, guide and guardrail its behaviours. The more the robot \emph{can} do, the more it has to be constrained.

One can try to save the singular reward principle and argue that appropriate constraints can be incorporated in the environment or in the reward function. Such enhancement of the kitchen or the concept of cleanliness is indeed possible. But it also calls for an exogenous intelligent human being whose duty is to impose these constraints on the robot -- so that it acquires the appropriate, rather than inappropriate, set of perceptual abilities and behaviours. In order to introduce a new type of utensils, a new kind of domestic animal (e.g. a parakeet), or a toddler in the family -- requires either making sure that the existing set of constraints works in the new setting or, if it does not, producing new constraints which are incorporated in the robot's perceptual, cognitive and action machine.

The kitchen robot therefore must know \emph{beforehand} what it must not do. It is absurd to hope that the robot learns what is bad, e.g. the constraint "thou shall not kill", in a sequence of trials and errors, as the authors imply. There is simply no practical way to ensure that the robot learns not to break glass and crockery and not to bend cutlery in multiple attempts with occasional human feedback. The agents must possess the concepts of what behaviour is prohibited in a particular setting, and what's bad in itself, absolutely. This knowledge is exogenous to the robot's task and must be imposed externally.

Next, suppose the robot knows what it \emph{must not} do. Then, from the get go, how does it know what it \emph{must do}, what is good to do? 

"Cleanliness", as occasionally evaluated by a human being, includes a number of very specific elements: clean floors, the oven, and the working surfaces, spotless wine glasses and utensils, the absence of smells, etc. The corresponding examples must be explicitly presented to a robot for the state-action-reward learning process to be efficient. Perhaps a kitchen robot can indeed learn what a spotless glass is by trial and error - by chance, after numerous iterations, requiring the constant presence of a human evaluator. Presentation of relevant samples of good behaviours by a human being can increase the efficiency of learning but it also automatically introduces human intelligence in the learning process. 

Save for trivial examples, the rewards of a real world agent are always multi-dimensional and the objectives are often in conflict. A kitchen robot must be aware of the trade-offs between conflicting objectives when time is limited; a social network algorithm must know about the trade-off between the revenue and the quality of the material presented to the user. The resolution of the trade-offs is non-trivial, specific to the problem, and is likely exogenous to the agent.

\centerline{***}

As every practitioner knows, shaping an agent's behaviour is difficult. 

On the one hand, there is always a possibility that the constraints are too tight and the actual behaviour of the agent is going to be driven most of the time by a set of human-imposed constraints. On the other hand, insufficient constraints, or the insufficiently comprehensive reward structure creates a risk of unwanted -- destructive, disorderly, unpredictable or less explainable --  behaviours. Looking at the robot example it is now evident that more complexity and nuance are required in understanding intelligence, natural and artificial, than is permitted by the "reward is enough" principle. 

Let us now explore the intellectual origins of "Reward is enough" to understand why and where the previous attempts of \emph{reductio ad lucrum } as an overarching principle have failed.

\section{Reward is not enough for understanding intelligent economic activity}

The concept of a human being pursuing wealth "who is capable of judging the comparative efficacy of means for obtaining that end" goes back to John Stuart Mill. He circumscribes the area of the concept's applicability, which I find is one of the signatures of a solid hypothesis. Political economy, as Mill explains, "predicts only such of the phenomena of the social state as take place in consequence of the pursuit of wealth" \cite{Mill1967}. But even in this limited capacity, it turns out, the principle fails to explain observable behaviours.

The concept of economic actors rationally maximizing their wealth found its apex in the Arrow-Debreu-McKenzie model of a closed economy in equilibrium.  Reward maximisation has also been successfully applied to study the political behaviour of voters, politicians and bureaucrats, and to describe rent-seeking -- entrepreneurs seeking rents and politicians and bureaucrats generating rents via regulation in a competitive process) \cite{Krueger1974}, \cite{Tullock}, and in constitutional decision making, namely, the choice of rules within which the activities of politics play themselves out \cite{Buchanan}, and others.

While the {\em Homo economicus} model is a useful mental construct which could be enhanced and applied with more or less success to a variety of observable human goal-seeking activities, its limitations are almost self-evident. The two examples demonstrate it clearly.

\subsection{The problem of innovative entrepreneurship} 
Despite the existence of productive innovators and entrepreneurs who are sometimes rewarded with wealth, recognition, and influence for their activities, the {\em expected} reward flowing to a {\em typical} innovator and his or her family is likely negligible or, in relative terms, even mildly negative. The bulk of expected total rewards of productive innovation is most often captured by the rest of society.

While it's possible to enhance the innovator's expected reward with such ephemeral virtues as "calling" and "personal fulfilment" in order to make the expected reward more positive, it is by no means self-evident why "calling" and "fulfilment" are available to but a small fraction of individuals.  To maintain the centrality of rewards in the analysis of innovation and entrepreneurship it is necessary to include the cultural and institutional factors.  How these factors contribute to entrepreneurs' and innovators' rewards given their individual trajectories, preferences, and constraints can by no means be described by a singular reward. 

Capturing all sources of an entrepreneur's motivation, both intrinsic or extrinsic, in a singular reward function is an impossible task. Some are motivated more by the aspirations of financial success, others by affiliation or self-acceptance \cite{Grouzet_2005}. The goals and the very content of entrepreneurship differ in different cultures. 

Innovative entrepreneurship is a long game one typically commits to play. It has been well understood for almost half a century -- at least in the context of economic science -- that commitment to a particular policy or action is explicitly and willfully sub-optimal \cite{sen1977rational} and cannot be reduced to reward optimization, incentives, or exogenous constraints. 
 
One can argue in favour of the possibility of enhanced or composite rewards. This argument, however, requires that the entire complexity of the environment in which entrepreneurs and innovators operate is incorporated into the innovator's reward structure. Under such circumstances, it seems impossible to identify exactly what the entrepreneurs' goals are, what exactly they maximize and, most importantly, how they know they actually maximize it.

\subsection{The problem  of crime}

Despite the overwhelming evidence that alternative activities are much more rewarding, individual and organized crime is endemic in some parts of a modern developed society. From a {\em Homo economicus} perspective a criminal is engaging in the most attractive (i.e. most rewarding) behaviour available. It is not, however, clear whether this is indeed the case --   given the massive evidence to the contrary, e.g. the size of the cumulative lifetime income of a typical criminal. 

The behaviour of a criminal is by no means shaped by the expected reward.

The availability of alternatives for the agent choosing between a criminal and a non-criminal activity is highly subjective: it does depend on the environment and the individual, his or her worldview, particularly on the ability to evaluate local costs and rewards, risks and uncertainties associated with each activity, and account for possible constraints. Even if a more rewarding alternative is made available and is known to a prospective criminal, the agent who is incentivised by a singular reward would have difficulties navigating all uncertainties and constraints to prefer a non-criminal activity and adjust his or her policy. 

The dimensionality of the problem explodes once we realize that none of the input variables to the societal and individual loss functions is known with certainty and that their shape is generally unknown and is most likely not static and depends on many exogenous factors.

The logic of "Reward is enough" leads us to believe that the availability of a singular reward, e.g. better expected lifetime/cumulative income, should help an individual avoid criminal behaviours. That this is not the case is a repudiation of the singular reward principle.

\subsection{Summary of reward maximisation in economics}

An attempt to approach both entrepreneurship/innovation and crime from the reward-maximizing goal-seeking standpoint suffers from serious logical flaws. 
    
{\em First}, in order for an individual reward-maximising agent to rationally engage in entrepreneurship, innovation, or crime, the corresponding reward function has to be computable (on a time scale relevant to the problem) by the agent {\em beforehand} given the state of the environment, knowledge of constraints and preferences of the agent, etc. 

It is almost self-evident that this is not the case -- aside from rare, trivial situations. Entrepreneurship, innovation, and crime are heavily affected by uncertainties which are often difficult or impossible to quantify or specify. They all require a rational or irrational commitment to chosen policies, often \emph{despite} information flowing from the environment and \emph{despite} the accumulated reward. 
    
{\em Second}, in order to learn the reward or loss function empirically (as an on-policy learning agent), an individual must engage in the corresponding activities multiple times. It is possible to imagine the accumulation of a moderate number of cases during the lifetime of an agent. It does seem to be possible for small, frequent innovations, replicative or subsistence entrepreneurship, and petty crimes. It is not clear, however, whether such an agent has enough opportunities to reliably estimate the expected reward, tail risks, and whether the agent is capable of separating the effect of endogenous inputs (the agent's policy or skill) from exogenous factors (luck or lack thereof).

\begin{figure*}[htb]
    \centering
    \includegraphics[width = 0.9\textwidth]{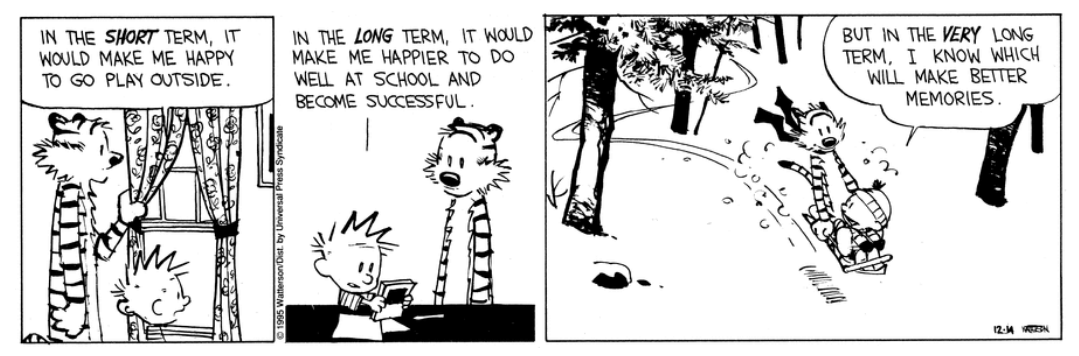}
    \caption{Intertemporal reasoning and decision making with robust heuristics and mental shortcuts. CALVIN AND HOBBES © Watterson. }
    \label{fig:Calvin}
\end{figure*}

In the absence of an easily computable cost and reward function intelligent human beings must instead rely on robust heuristics, mental shortcuts, norms, customs and habits, as cleverly shown in Figure \ref{fig:Calvin}.

Despite its flaws and limitations, the concept of \emph{Homo economicus} (as well as, more generally, of agents maximising their wealth) is a useful mental tool to reason about some behaviours. And keeping in mind its limitations, it does not seem a huge stretch to try to use it even in a multi-agent setting. Some productive and interesting examples include
\begin{itemize}
    \item Analysis of property rights in oligarchic vs democratic societies in \cite{Acemoglu_10.2307/40005150}. The model comprises a continuum of zero intelligence agents who maximise their discounted consumption flow and engage in entrepreneurship, both economic and political. The model produces quite reasonable conclusions regarding the advantages and disadvantages of both political systems.  
    \item Analysis of zero intelligence agents operating in continuous double auction markets (e.g. stock market, Forex, crypto exchanges) and optimising their profit -- e.g. \cite{Cliff_ZIP60} and references therein. The model produces believable behaviours and demonstrates its usefulness: the influence of the rules of the game on the agents' behaviour can be analysed and the rules themselves be improved. 
\end{itemize}

It is curious that both these examples clearly demonstrate that quite complex social phenomena we tend to associate with intelligent activities can be modelled using zero- or low-intelligence  agents. It looks like the entity which is actually responsible for the emergent complex phenomena is not the community of agents as such but the set of exogenous rules of the game.

It is encouraging that such simple models of reward maximising agents can reproduce a limited set of \emph{stylized facts} for systems of limited complexity. It is also clear that what these models reveal is quite narrow. They tell us almost nothing about how real intelligent agents act in real environments.

A real economy is never in equilibrium and its future trajectories are uncertain. Some of these uncertainties are coming from the environment, but mostly they are due to the economy's internal complexity. Real economic agents have different views of the future, different preferences, and pursue different objectives. The real economic activity of such heterogeneous agents is an emergent phenomenon. In a real economy, each economic agent tries to find a sustainable specialisation according to the agent's beliefs, preferences, values, and circumstances, and is engaged in trade with other agents. Trade is made possible by agents having different beliefs and different specialisation. Fluid patterns of (temporary) sustainable specialisation and trade give rise to the modern economy \cite{kling2016specialization}. Reward maximisation plays a limited role, but costs that enter into the decision of each agent are mostly subjective and the future rewards are always uncertain.

In a real economy, even the rules of the game are not given and therefore cannot be assumed to be the inputs of the reward function. There is a significant, varying from one society to another, informal component in the real economy's rules of the game.  "One does not negotiate the rules of chess. But informality is continuously negotiated... [Rules]  are not about regulating relations between pre-existing people and objects. They are about creating entirely new power relationships between people, and then continually negotiating about them." \cite{mccloskey_2016}.

Simplification is inevitable in human cognition and stylized representations are common in the branches of human inquiry dealing with complex systems. Mapping from observations to various bounded representations has been the cornerstone of human cognition since the earliest days of civilization. Behind each successful representation mapping, there is always a pragmatic attempt to preserve and make use of bits and pieces of information that matter in a particular setting and for specific purposes. This pragmatism is well expressed by the dictum: "A map \emph{is not} the territory it represents, but, if correct, it has a \emph{similar structure} to the territory, which accounts for its usefulness" \cite{korzybski1933science}. 

It is its lack of pragmatic usefulness that makes the lens of reward maximisation a poor tool for conceptualizing the complexities of real economic behaviours of intelligent human beings. Reward maximisation is an incomplete framework and cannot describe other known individual and social behaviours, as I will show now. 

\section{Reward is not enough for human behaviour} \label{sec:behaviourism}
 \cite{SILVER2021103535} insist that the reward-maximization is not behaviourism:
\begin{quote}
     [T]he reinforcement learning problem differs from behaviourism in allowing an agent to construct and use an internal state. 
\end{quote}
The opposite is true: the reinforcement learning paradigm as circumscribed by \cite{SILVER2021103535} \emph{is} essentially and substantially covered by behaviourism.

Behaviourism precedes the reinforcement learning paradigm by almost a century \cite{sep-behaviorism}.  Behaviourism's concepts are mapped to the concepts of reward maximisation: e.g. "stimuli" and "responses" directly corresponding to the "states" and "actions", and the positive and negative reinforcement and the positive and negative punishment, corresponding to rewards in the reinforcement learning paradigm. Such conceptualisation makes it possible to express the adjustment of the "operant behaviour" - i.e. "operant conditioning" of behaviourism -- precisely in terms of the reinforcement learning paradigm. Except for the abundance of computational techniques and clever demonstrations and limited practical successes, reward maximisation and reinforcement learning do not expand the behaviourism framework: "reward is enough" is a modern, more limited and focused, reincarnation of radical behaviourism. 

Even the most radical version of behaviourism does not reject the possibility that the agent constructs and uses its internal state. In relation to the internal states of the agents, behaviourism simply states that the individual history of the agent, the sequence of stimuli, responses and reinforcement, plus its origins and its genetics -- plays a primary role, whereas the psychological state of the agent is either secondary or irrelevant \cite{Skinner_AboutBehaviorism}. 
\begin{quote}
 [W]hat is felt or introspectively observed is not some nonphysical world of consciousness, mind, or mental life but the observer's own body...
    An organism behaves as it does
because of its current structure, but most of this is out of reach of introspection. At the moment we must content ourselves, as the methodological behaviourist insists, with a person's genetic and environmental histories.
What are introspectively observed are certain collateral products of those
histories. 

The environment made its first great contribution during the evolution
of the species, but it exerts a different kind of effect during the lifetime of
the individual, and the combination of the two effects is the behaviour we
observe at any given time. Any available information about either contribution helps in the prediction and control of human behaviour and in its
interpretation in daily life. To the extent that either can be changed,
behaviour can be changed. \cite{Skinner_AboutBehaviorism}
\end{quote}

From the practical perspective, the effect of the environment and reinforcement on the agent and the malleability of its policies are central to behaviourism just as well as they are central to the reinforcement learning paradigm. Positive and negative reinforcements and positive and negative punishments induce directed changes in the agent's environment. Repeated reinforcement gradually changes the agent's policies and thus leads to the desired change in its behaviour.

Many initial and subsequent behaviourist insights were obtained experimentally and therefore qualified as scientifically objective.  It would be hard to argue against objectively obtained results. Indeed, the use of rewards and punishments to train domesticated animals to perform certain tasks has been known to produce reliable results for millennia. From this perspective, the excellent demonstration of the reinforcement learning approach in the realm of artificial agents is just another practical application of behaviourism.

Problems begin when an experimentally confirmed theory is extended beyond its area of applicability. In "Selection by consequence" B.F. Skinner \cite{Skinner_Science_10.2307/1686399} expresses natural selection as selection by consequence (= reinforcement) and extends it to psychological phenomena, social, and cultural behaviours, and uses it to explain the emergence of language-mediated communication etc. 
\begin{quote}
By behaving verbally people cooperate more successfully in common ventures. By taking advice, heeding warnings, following instructions, and observing rules, they profit from what others have already learned... A culture evolves when practices originating in this way contribute to the success of the practising group in solving its problems. \cite{Skinner_Science_10.2307/1686399}
\end{quote}
Parallels with "Reward is enough" are evident.  
\begin{quote}
According to our hypothesis, the ability of language in its full richness, including all of these broader abilities, arises from the pursuit of reward. It is an instance of an agent's ability to produce complex sequences of actions (e.g. uttering sentences) based on complex sequences of observations (e.g. receiving sentences) in order to influence other agents in the environment (cf. discussion of social intelligence above) and accumulate greater reward...
\end{quote}
For behaviourists "operant conditioning" (= policy optimization) in response to "selection by consequence" (= reinforcement, reward maximisation) can explain arbitrarily complex behaviours, too:
\begin{quote}
    The immediacy of operant conditioning has certain practical advantages. For example, when a currently adaptive feature is presumably too complex to have occurred in its present form as a single variation, it is usually explained as the product of a sequence of simpler variations, each with its own survival value. It is standard practice in evolutionary theory to look for such sequences, and anthropologists and historians have reconstructed the stages through which moral and ethical codes, art, music, literature, science, technology, and so on, have presumably evolved. A complex operant, however, can actually be "shaped through successive approximation" by arranging a graded series of contingencies of reinforcement \cite{Skinner_AboutBehaviorism}
\end{quote}
This view is paralleled by \cite{SILVER2021103535}, too. 

In attempting to turn a successful practical method of limited applicability into an overarching principle the ambition of the authors of "Reward is enough" almost matches that of behaviourists. The overlap between the two programs is evident. Both programs propose a rigid relationship between the environment and the agent. This relationship is meant to explain and predict the complexities of the behaviour of living organisms and artificial agents. Both view reinforcement, selection, and reward optimisation as overarching factors capable of shaping complex and intelligent activities. Behaviourism does "not deny the possibility of self-observation or self-knowledge or its possible usefulness" \cite{Skinner_AboutBehaviorism}. Behaviourism is just very sceptical about the possibility of full observability of inner states.

Behaviourism attractively differs from the \emph{Homo economicus} framework in one important aspect: it explicitly recognizes the possibility of multiplicity and multidimensionality of stimuli and the corresponding rewards. 

But just like the \emph{Homo economicus} model, behaviourism turns out to be not specific enough to describe intelligence. It is now recognized that it, too, is an incomplete model:
\begin{quote}
    Why has the influence of behaviourism declined? The deepest and most complex reason for behaviourism’s decline in influence is its commitment to the thesis that behaviour can be explained without reference to non-behavioural and inner mental (cognitive, representational, or interpretative) activity. 

    Skinner’s vantage point on or special contribution to behaviourism mates the science of behaviour with the language of organism/environment interactions. But we humans don’t just run and mate and walk and eat in this or that environment. We think, classify, analyze, imagine, and theorize. In addition to our outer behaviour, we have highly complex inner lives, wherein we are active, often imaginatively, in our heads, all the while often remaining as stuck as posts, as still as stones \cite{sep-behaviorism}.
\end{quote}

Indeed, some fairly complex behaviours of living organisms and artificial agents can be explained as being conditioned by stimuli and shaped by reinforcement through the individual agent's trajectory or genetically. One of the simplest organisms, \emph{Amoeba proteus}, is known to be capable of conditioned locomotion and, possibly, even a short-term associative memory \cite{Amoeba_Proteus_2019}.  

The behaviour of higher organisms and human intelligence are different. "Reward is enough" is an incomplete framework and cannot fully describe and explain it.

\section{Reward is not enough for perception}


Abilities to perceive are a subset of abilities needed for the survival of an organism  or for the successful operation of an artificial agent. All abilities are costly to maintain, they compete for the limited resources of the organism. 

Each sensory ability along with its particular implementation represents an evolutionary trade-off. The evidence of the cost-function trade-offs is abound: 
\begin{enumerate}
    \item Despite potentially diminished rewards from the reduction in perception abilities some organisms in the process of evolution do lose some of their perceptional abilities -- apparently to achieve better performance in others.
    \item The complexities of the sensory apparatus of various animals depend on their evolutionary trajectories as well as the specifics of their environments or rewards: nocturnal birds have more rod cells than colour sensitive cone cells, and birds can see the UV light. Despite the multiple potential rewards associated with the ability to see the UV light, humans cannot see it.
\end{enumerate}
Perception in higher animals is mediated by a variety of coding schemes. A particular scheme, e.g. population coding, better serves some functions and is less capable of serving others. It does not seem possible to derive a particular scheme and its specific implementation from the singular reward maximisation only. 

The reinforcement learning paradigm suggests that an artificial agent -- fed with a wide array of sensory inputs -- eventually develops a set of \emph{relevant} perceptional skills necessary to maximise a reward in a particular setting. Toy examples suggest that a well-calibrated agent can indeed distinguish between more informative and less informative inputs. However, achieving sensory efficiency in the presence of trade-offs requires more than just a proportional response to rewards.

In practice, the trade-off between the artificial agent's function and its complexity does require informative and efficient inputs. Often they are crafted by human engineers. Trial and error and state-action-reward process and occasional human feedback cannot help an artificial agent invent a lidar sensor. Most of the information coming into the cognitive machine of a self-driving car is designed by a human intellect.

The existence of trade-offs between the costs associated with perception and the potential rewards is fatal to the maximisation of a singular reward principle.  Costs associated with various perceptual abilities are environment- and problem-dependent. A web-building spider has relatively poor vision but is endowed with excellent sensitivity to touch and vibration. An improvement in the spider's eyesight theoretically brings more rewards (there are spiders with excellent night vision, for example), but good eyesight is costly to maintain \emph{in the particular environment} where the web-building spider operates. Optimisation of the combined reward-cost trade-off function, if it indeed can be constructed, is the equivalent of a "specialized problem formulation" -- which is needed for the agent's success.

Costs and rewards associated with different perceptual skills are incommensurable. In order to construct the trade-off function for an artificial agent the multiple components of costs and rewards need to be measured in the same units. And the trade-off is also environment- and problem-specific. A singular reward maximisation is not enough to drive the development of perceptual abilities.

\subsection{Coding schemes}\label{sec:coding}

Coding theory represents another good example of the incompleteness of the reinforcement learning and reward maximisation approach. 

Efficient data or signal transmission can be understood as part of the problem of efficient perception. There are several ways in which living organisms optimise internal signal transmission. Population coding is how photoreceptors, taste cells, or olfactory cells transmit their information to the brain \cite{bear2020neuroscience}. These sensor cells are tuned broadly, so that there is often a substantial overlap between the sensitivity of different cells of the same type\footnote{Spectral sensitivities of the red and the green photoreceptors are $575\pm100$ nm and $535\pm100$ nm, respectively. }.  The brain receives partially complementary and partially overlapping signals. The scheme provides the double benefit of signal discrimination and error correction at the cost of maintenance of the multiple transmission channels. 

In population coding the same signal can be sent via $N$ different channels to the brain where it is reassembled. From the information-theoretic perspective such repetition code is the simplest possible approach to noise reduction and is reasonably effective: the probability of error falls exponentially with each repetition $\epsilon \propto f^N$, where $f<1$ is roughly the probability of an error in each transmission. 

The efficiency of repetition codes is low: the same signal is transmitted multiple times to correct for an error in a part of a signal. The cost of transmission grows linearly with each repetition. As a function of repetitions the trade-off function $R(N)$ definitely has a maximum:
\begin{equation*}
    R(N) \propto - \alpha f^N - \beta N  
\end{equation*}

But the location of this maximum depends on the agent's circumstances expressed by $\alpha, \beta$. There is no unique, objective way to express the cost of transmission and the benefit associated with the error reduction in the same measurement units. 

One can indeed imagine an evolutionary process that leads to the maximization of $R(N) $: e.g. black and white vision, one extra colour cell type, two overlapping colour sensitive cells, infrared, ultraviolet, etc. until it settles on a most beneficial combination for a particular species in particular circumstances, an equivalent of a specialised problem formulation, the opposite of a singular reward maximisation.

What one cannot imagine is a trade-off function optimisation leading to the invention of block codes \cite{mackay2003information}. 

Block coding is super efficient in ensuring that massive amounts of information are stored and transmitted error-free. A typical block code stores information in blocks. Each data block is accompanied by a smaller block carrying a summary of the data block such as its parity or a check sum. The summary block is sufficiently rich to permit, to an extent, the recovery of the data block if it is corrupted. This scheme entails very efficient error reduction at a very modest cost of transmission of just a little bit of extra data. 

The difference between block codes and repetition codes is not in degree but in nature. While the latter are \emph{optimizations} achievable potentially through a trade-off optimisation, the former are \emph{inventions}. Inventions are human creations. 

Efficient coding schemes in modern data storage and transmission do produce truly enormous rewards for all of us. But they cannot be re-created by an agent, however complex, trained to optimise singular rewards.

"Reward is enough" is a poor lens to understand perception.

\section{Reward is not enough for knowledge}


 \cite{SILVER2021103535} define knowledge in a somewhat narrow sense, as "information that is internal to the agent". The vague definition is self-evidently unsatisfactory within the authors' own framework: the agent can be full of false or irrelevant internal information and can have true or false beliefs unjustified by any rewards, and yet can act reasonably efficiently  -- e.g. due to exogenous constraints. 

One may attempt to save the "reward is enough" principle by adding "true" and "relevant" to the "information that is internal to the agent". What is true and what is relevant is often a matter of degree rather than of kind, and, more importantly, is often defined with respect to a particular function or skill and the environment in which the agent operates. This fact immediately makes the agent's knowledge dependent on the specialised problem formulation. 

\subsection{Uses of imitation, learning and knowledge  in artificial agents}

A real world AI agent typically operates in a highly constraining embedding. Compliance with constraints, that is, exogenous conditions which permit some actions and prohibit others in particular states of the world, naturally interferes with the agent's skills learned by imitation or knowledge accumulation, and can even make learned behaviours irrelevant: a heavily constrained agent can be driven entirely or most of the time by constraints or rules of the game. 

To save the "reward is enough" principle one could argue that the agent's constraints can be incorporated into its innate knowledge. This, again, destroys the singular reward principle: constraints are \emph{always} specific to the problem and the function of the agent. An agent that follows constraints, as I demonstrated earlier in the paper, follows the specialised problem formulation rather than maximises a singular reward.

 \cite{SILVER2021103535} introduce a relevant real world example of an agent operating in a changing environment where it needs to find a balance between innate and learned knowledge, but they do not go far enough. The balance between the learned and the imposed knowledge in a changing environment is one of the central problems in the development of real world artificial agents.  Knowledge acquisition through imitation or trial and error and observation has its costs: internalising the agent's observables into knowledge requires time and computing capacity. In a quickly changing environment, there is a limit on the rate of change with which the agent can internalize observations \emph{and} successfully act on the newly acquired knowledge. This limit depends on the complexity of the cognitive machine of the agent, its perceptual abilities, and the extent and dimensionality and granularity of its decision space. The trade-off between the costs of knowledge acquisition and the benefits of its use generally cannot be decided or optimized by the agent. In the absence of other inputs about the environment, the relative importance of innate vs learned knowledge must be supplied externally or heuristically and is inevitably dependent on the agent's purpose, and its function: it is generally defined by a problem formulation specific to the agent's objective and its environment. 

\section{Evolution}

For \cite{SILVER2021103535} the mere existence of an organism is the confirmation of a singular reward maximisation mechanism:
\begin{quote}
     Evolution by natural selection can be understood at an abstract level as maximising fitness, as measured by individual reproductive success, optimised by a population-based mechanism such as mutation and crossover. In our framework, reproductive success can be seen as one possible reward signal that has driven the emergence of natural intelligence. 
\end{quote}

The reinforcement learning view of evolution parallels behaviourism's "selection by consequence" \cite{Skinner_Science_10.2307/1686399}. Both are oversimplifications. 

Using reproductive success (or any other singular reward) as the driver of evolution toward natural intelligence is problematic.  Plants, by their individual reproductive success,  are the most successful type of organism on Earth by their biomass, bacteria being the distant second \cite{Bar-On6506}. 
Reproduction of higher primates including humans is another challenge for considering reproductive success as a signal driving evolution. Higher primates and humans mostly produce singletons or twins rather than litters. The risky and costly evolutionary change \cite{heying2021hunter} is a step back from the litter-based reproduction.  The group-level reproductive success is not self-evident as well: it took several million years for humans to outnumber other mammals except perhaps bats and rodents who occupy essentially the same natural habitat.

The modern understanding of evolution as variability-driven exploration in the space of traits and resources offers a broader and richer framework of life on Earth. It does not preclude local optimality of any traits, including the reproductive success of an organism,  but does not make it a requirement or a goal.

\begin{itemize}
    \item Ever present genomic and developmental variability (including horizontal gene transfer) produces changes in traits of living organisms  (e.g. structures and functions and behaviours)
    \item Trade-offs are inherent: an advancement in one trait is often accompanied by a decline in others 
    \item Genomic and developmental variability is mostly neutral with respect to the overall fitness metrics; it is constrained by the removal of harmful changes \cite{koonin2009origin}, \cite{Koonin2009DarwinianEI} 
    \item Neutral exploration in the space of traits due to genomic and developmental variability is accompanied by the exploration in the space of resources (e.g. nutrients, including those produced by other organisms, energy, physical space)
    \item Exploration in the space of resources inevitably leads to interaction, competition and collaboration with other organisms operating in the same space
    \item The result of exploration in the space of traits and resources continuously creates and destroys opportunities for individual organisms and groups; it is non-directional, does not lead to any evolutionary progress, does not necessarily produce more complex organisms, is often gradual, but sometimes occurs in leaps due to population bottlenecks or unique events
    \item Observed biodiversity is a dynamic function of available resources and historical trajectories of organisms in the space of traits and resources
\end{itemize}

Reproduction is of course necessary to engage the mechanisms of exploration in the space of traits and resources but it is by no means the objective of the process of evolution. 

One could argue that the above evolutionary framework can still be thought of as a variant of the "reward is enough" hypothesis with resource consumption maximisation replacing reproductive success.  This view must be rejected: resources are functions of specific circumstances and are always multifaceted -- the singular reward principle does not apply. 

The remaining possibility is to apply reproductive success as the driving signal of evolution on the genomic level, following the line of thought popularized in "The Selfish Gene" \cite{Riddley_RePEc:nat:nature:v:529:y:2016:i:7587:d:10.1038_529462a}. 

This reasoning suffers from the same flaws but on a more granular level. Genes are just as fluid, as dynamic, as modular, and as responsive to their environment as macro-organisms and micro-organisms \cite{Comfort_RePEc:nat:nature:v:525:y:2015:i:7568:d:10.1038_525184a}: they compete and collaborate, interact with other genes, and thus maintain their diversity. The apparent excess genetic material in organisms supports the idea of exploration in the space of traits and resources -- by the organism, and by the gene pool.

On both levels -- organism and genetic -- evolution can be viewed as an emergent property of interacting biological systems rather than a directional change driven by a one-factor reproductive optimization. "Reward is enough" cannot describe observable facts of evolution.

\section{Language, culture, other social intelligent behaviours}

In section \ref{sec:behaviourism} we already mentioned multiple overlaps between the "reward is enough" hypothesis and behaviourism in their treatment of language, culture and social behaviours. 

Behaviourism's approach seems slightly more attractive and broader than the reward maximisation paradigm: while it keeps the principle of selection by consequence (= reward) as the driving signal, it takes a more explicit evolutionary view of culture, as evidenced by this quote:
\begin{quote}
    ...human behavior is the joint product of (i) the contingencies of survival responsible for the natural selection of the species and (ii) the contingencies of reinforcement responsible for the repertoires acquired by its members, including (iii) the special contingencies maintained by an evolved social environment. (Ultimately, of course, it is all a matter of natural selection, since operant conditioning is an evolved process, of which cultural practices are special applications.) \cite{Skinner_Science_10.2307/1686399}
\end{quote}


Formal equilibrium solutions of game theory cited by \cite{SILVER2021103535} are not applicable to real languages and cultures: social behaviours are never in equilibrium. Just like the real world economics, languages and cultures are emergent phenomena. 

Language is an example of an emerging social behaviour that demonstrably does not fit into the "reward is enough" principle. 

Discussing the emergence of language, \cite{SILVER2021103535} derive it through a multi-agent reinforcement learning process. Language 
\begin{quote}
...is an instance of an agent's ability to produce complex sequences of actions (e.g. uttering sentences) based on complex sequences of observations (e.g. receiving sentences) in order to influence other agents in the environment (cf. discussion of social intelligence above) and accumulate greater reward    
\end{quote}

This could partially explain some of the most primitive uses of language. But it fails to explain the mechanism of tactical deception by capuchin monkeys who use alarm calls ('cry wolf') in a functionally deceptive manner to usurp food resources. Deception could bring short-term rewards to a social agent engaged in deceptive behaviour, but the agent itself and the commune also incur costs. The probable rewards flowing to the individual and the costs incurred by the commune are impossible to reconcile or even measure. A complex and rich structure of costs, benefits, and constraints actually defines the specific problem of the use of deception in social behaviours, and the specific mechanisms necessary to resolve the emerging issues -- a singular reward is not enough.

The reward-centric hypothesis cannot explain, contrary to the assertion by \cite{SILVER2021103535}, higher uses of language such as literature, art, theatre, philosophy, religion, and scientific communication. The evolution of the 'cry wolf' deceptive signal into literature does not necessarily involve any immediate or distant rewards and can even be extremely costly for the agent who comes up with such an innovation:
\begin{quote}
    Literature was born not the day when a boy crying wolf, wolf came running out of the Neanderthal valley with a big gray wolf at his heels: literature was born on the day when a boy came crying wolf, wolf and there was no wolf behind him. That the poor little fellow because he lied too often was finally eaten up by a real beast is quite incidental. But here is what is important. Between the wolf in the tall grass and the wolf in the tall story there is a shimmering go-between. That go-between, that prism, is the art of literature \cite{nabokov2017lectures}.
\end{quote}

One could argue that the emergence of higher uses of language can ultimately be traced to some specific rewards: e.g. literature, art, and theatre facilitate social cohesion; philosophy and religion define what is good and what is not; and scientific communication facilitates the exchange of potentially reward-bearing ideas. But if these rewards are specific to each activity then the "specialised problem formulation" is a better lens through which we can view these higher uses of language than the "reward is enough" principle. Second, neither social cohesion, or systematic ethical structures, or the specific forms of exchange of ideas bring unequivocal and unconditional rewards: social cohesion sometimes contributes to horrible wars, ethical orthodoxy and institutionalised forms of scientific exchange sometimes impede society's ability to adapt and produce new ideas, etc. 

An alternative lens sees social behaviours as tools.

\begin{quote}
    ... all human languages are tools. Tools to solve the twin problem of communication and social cohesion. Tools shaped by the distinctive pressures of their cultural niches --- pressures that include cultural values and history and which in many cases account for similarities and differences between languages \cite{everett2012language}.
\end{quote}

There is plenty of evidence that the use of language and its characteristics are mediated by culture, and that a particular culture, understood as socially transmitted cognitive and behavioural patterns, is affected by that culture's language and its structure (see, e.g. \cite{boroditsky2001does}, \cite{boroditsky2003sex}, \cite{Hu_He_Colour_10.3389/fpsyg.2019.00551}). Language, through its expressive devices, can mediate our decision  making \cite{thibodeau2011metaphors}. A singular reward maximisation cannot explain the well studied empirical fact that "categories in language
can affect the performance of basic perceptual colour discrimination
tasks" \cite{winawer2007russian}.  "Tools shaped by distinctive pressures", rather than the singular reward hypothesis, is a better framework for the interplay between language, culture, and the corresponding constraints and trade-offs.

Continuing this line of thought: all social behaviours can be viewed as tools which solve multiple problems and which are shaped by their distinctive social pressures. Languages and cultures and other social behaviours are stable in some respects and are changing in other respects - they are shaped by the multifaceted pressures of the embedding in which they operate.

Approaching social behaviours from this perspective rules out the singular reward principle. These tools are just too complex themselves, and too intertwined between themselves, and the problems they solve are too complex and intertwined to fit into the singular reward maximisation paradigm. "Reward is enough" does not tell us why and how these social tools evolve, how they are shaped, and what explains the similarities and differences between them. 

\section{Summary of the critique of the "reward is enough" principle}

"Reward is enough" as a lens intended to help us understand intelligence and the associated phenomena essentially overlap with its failed intellectual predecessors -- behaviourism and the \emph{Homo economicus} model. The singular reward principle is too narrow to reason about evolution, social behaviours, culture, language, knowledge, learning, imitation, and perception.

\section{What is intelligence -- from a practitioner's point of view}

A bee struggling to get out of a room \emph{through} a glass window pane exercises a typical state-action-reward  behaviour which is likely hardcoded in its tiny brain. This policy is unlikely optimized to produce a specific action or an action pattern in response to the specific state in order to collect a reward. Rather, it is a generic policy that is part of a more high-level policy that guides the bee to explore the world: the anisotropy of light guides the bee away from the already explored dark place. While the generic policy is not optimized for the specific situation, it sort of works: by sheer luck, some bees do manage to escape before they die of starvation.  What is clear is that the bee's state-action-possible reward policy is not the behaviour of an intelligent being. 

Bee's more specialised and complex activities, e.g. those involved in the construction of a honeycomb, may resemble the work of an engineer or an architect. It has been demonstrated that the regular structure of a honeycomb is a product of the activity of many individual bees who follow a simple built-in rule and use their body parts to measure distances \cite{Honeycomb_Nazzi_2016}. A cellular automaton following an externally imposed rule can create a structure resembling a honeycomb. The built-in rule is likely the result of exploration in the space of comb-building activities and can indeed be an optimal cost- and information-saving policy. It is very likely that a carefully constructed reward maximisation experiment can indeed yield an optimal honeycomb-producing rule. But the remarkable regularity and efficiency of a honeycomb is not the product of an intellect.

\subsection{Reflexive behaviours}

Both the light-triggered escape policy and the honeycomb policy belong to the class of \emph{reflexive behaviours}. Innate behaviours such as reflexes and fixed action patterns supply quick and reliable responses in standard situations. Most of the impressive demonstrations of the singular reward maximisation and the reinforcement learning methodology can be classified as models reproducing reflexive behaviours or fixed action patterns. 

Shaped and hardcoded by evolutionary forces of behavioural variability and exploration,  reflexes and fixed action patterns of natural agents can also be modulated, to a degree, to accommodate new circumstances. Reflexive behaviours of artificial agents can also be modulated in response to newly acquired observations. What portion of the innate knowledge should be discarded to modulate the agent's behaviour depends on the balance of many factors including the cost of internalizing the new observations, the likely benefit of the modulated behaviour, and the rate of the environmental change. 

\subsection{Associative learning and habituation}

Behaviours shaped by associative learning also fall into the reinforcement learning paradigm. However, associative learning is not specific to intelligent behaviour. It is a universal adaptive mechanism shared by almost all natural organisms \cite{vanDuijn_bio_cognition} and primitive artificial agents. A set of matchboxes can learn how to play a simple game of six pawns (Hexapawn \cite{gardner_10.2307/24937263}, \cite{gardner1986unexpected}) by associating a move in a particular state with the probability of being on a winning trajectory. The ability to play this miniature game is not evidence of intelligence. Arguably, the complexity of the game notwithstanding, AlphaGo and AlphaZero are only quantitatively different from Hexapawn.

Habituation -- i.e. learning not to respond to an irrelevant stimulus -- is more complex than associative learning.  Unlearning of this kind already requires special mechanisms which go beyond the standard reinforcement learning approach. Habituation is a risky strategy: a stimulus that seems irrelevant right now may become relevant again in the future. Natural agents develop the ability to habituate in response to a combination of environmental and internal pressures \cite{vanDuijn_bio_cognition}.   

An artificial agent capable of habituation should be made aware of why unlearning to respond to an irrelevant stimulus is good. The agent can weigh the relative cost of maintaining the ability to respond to an irrelevant stimulus versus the probable benefit of freeing the resources for different tasks or abilities. The balance of risks and benefits of habituation depends on many factors. The trade-offs of habituation are agent- and task- and environment-specific. They are resolved in a problem- and situation-specific way.

\subsection{Instincts and the innate ability for social learning}

Instincts are innate and often shape very complex activities, e.g. learning and mating behaviours. Activities associated with instincts can be morphed by adaptive learning, habituation, or even consciously -- by intelligent agents. 

Even primitive social organisms such as bumblebees can solve reasonably interesting puzzles and then teach the colony to modify the innate behaviours to accommodate the new skill. The problem of a trade-off between the limited cognitive capacity of an organism and the probable benefits of a new skill does not yield a single solution common to all agents, but is solved socially, through the diversification of cognitive skills: "there are huge inter-individual differences: most bees will require either step-wise training or the chance to observe a skilled conspecific to master the task on their own. A very small minority of individuals even solve the task by individual trial-and-error learning." \cite{Bees_cognition_Chittika_2017}. The trade-offs of a quick learner and those of a more cognitively stubborn individual do not optimise for a particular expected individual reward. Rather, they are there to solve a bigger problem - survival and adaptation of the whole colony.  How a particular distribution of cognitive preferences plays out depends on many factors and is specific to the colony's environment.

Artificial social agents tasked with complex instinctive behaviours that require online adaptation and learning can benefit from the diversification of their cognitive preferences: rather than all of them replicating the best local policy different agents can have different elasticities of their cognitive apparatus. Cognitive and behavioural diversification helps manage social risks.

\subsection{Individual foresight}

A 19th century thinker wrote: "what distinguishes the worst architect from the best of bees is that the architect builds the cell in his mind before he constructs it in wax" \cite{marx2004capital}. True intelligence is inseparable from the ability to imagine, to foresee, to prognosticate, to plan, to strategize. The ability to "build the cell in one's mind" almost always hinges on a limited view about the future. 

An intelligent agent differs from a non-intelligent agent (e.g. exhibiting reflexes, fixed pattern actions, and instincts) in its ability to perform a voluntary and planned action or a sequence of actions. A framework that explicitly takes into account the choice-centric character of an intelligent agent must explicitly recognize 
the fact that it cannot be truly represented "as an automation of reflexes, as a mind-machine, as a bundle of instincts, as a pawn of drives and reactions, as a mere product of instincts, heredity, and environment..." \cite{frankl2010doctor}. Perhaps a better framework for an intelligent agent is not states, actions, and rewards, but beliefs, desires, intentions (see e.g. \cite{rao91a},\cite{Vikhorev_BDI} and references therein), and the associated constraints and trade-offs.

An intelligent agent is capable of setting a goal for itself to achieve. The goal formulation can include reaching a certain state, e.g. arriving from Palo Alto to San Francisco by 17:00, or collecting uncertain rewards, e.g. the value of the agent's retirement fund.  Embarking on a path to reach a goal necessitates conscious evaluation and rejection of alternative future actions or inaction. To reach San Francisco from Palo Alto the agent can choose between the two alternative highway routes (HW 101 or 280), or try using local streets: there are advantages and disadvantages of either solution. To improve probable future outcomes of its retirement fund the agent can invest in stocks, government, corporate or municipal bonds etc -- with their respective advantages and disadvantages. The ability to choose is the characteristic of an intelligent agent. This choice first exists in the agent's mind, in a manufactured for this purpose model of reality, in a representation of reality produced by the agent's cognitive machine.

In order to be able to perform the act of a complex choice the agent needs a representation of reality that it can use to consider various scenarios of the future in relation to its actions. The mapping of reality to its representation is rarely exact, but it must preserve the essential structure of reality -- to be useful \cite{korzybski1933science}. In order to operate the representation of a complex environment an intelligent agent can create and use symbolic systems (e.g. math \cite{Wigner_https://doi.org/10.1002/cpa.3160130102}, myths \cite{campbell2017myths}). A useful representation of reality also includes constraints - hard and soft (e.g. rules of the game, norms, customs), as well as the agent's beliefs concerning what is good for the agent and what is not. An intelligent agent is self-aware.

Another important difference between a reflexive agent (e.g. the one trained with a standard reinforcement learning algorithm) and an intelligent agent is the ability of the intelligent agent to imagine and select a hypothetical future trajectory that \emph{is not represented by any training data}. 

An \emph{improvement} or optimisation is the result of examining past trajectories and selecting the actions that more often choose more optimal trajectories. This expectation maximisation strategy could work if the agent is given multiple chances to make the same decision in the same state of the world.  The strategy of using historical trajectories in decision making already becomes problematic when the agent encounters situations (= states of the world) where its confidence in the trade-offs associated with a particular action is low or when the agent makes a particular choice only once: expectations do not work. 

An \emph{invention}, on the other hand, such as the invention of block codes I discussed earlier in section \ref{sec:coding},  starts with an inventor-agent launching an avatar into the previously unseen territory. In an excellent essay on the problem of measuring natural and artificial intelligence Fran\c{c}ois Chollet writes:

\begin{quote}
    Task-specific performance is a perfectly appropriate and effective measure of success if and only if handling the task as initially specified is the end goal of the system – in other words, if our measure of performance captures exactly what we expect of the system. However, it is deficient if we need systems that can show autonomy in handling situations that the system creator did not plan for, that can dynamically adapt to changes in the task – or in the context of the task – without further human intervention, or that can be repurposed for other tasks.
\end{quote}

Precisely! 

Once the choice is made, commitment to a particular policy or trajectory is an attribute of a choosing agent. Commitment explains intelligent behaviours which exist seemingly against all odds and in contrast with any apparent reward maximisation -- e.g. chronic entrepreneurship, marriage, friendship etc.  

Commitment is especially important for social agents which I discuss below. 

\subsection{Social behaviour of intelligent agents}

Simulations of environments populated by artificial agents of various levels of intelligence have a long and rich history: from a team of intelligent agents performing specific tasks in a synthetic battlefield environment \cite{Intelligent_Agents_Battlefield_1997} to complex behaviours in electricity markets \cite{Multi-Agent-Electricity-2015} and financial markets \cite{belcak2020fast}. 

Many social agent simulations are mostly driven by externally imposed rules. Given a strict enough set of rules, even a commune of zero-intelligence agents produces believable \emph{macroscopic} characteristics of the target environment \cite{Realism_Multi_Agent_2011}. There are successful demonstrations of how relatively low complexity agents can learn to operate optimally in a stable rule-constrained environment -- see, e.g. \cite{spooner2020market}.  

Understanding how a group of agents can develop their own rules of the game is critical to the study of collective action. Some progress has been made in simulating the emergence of elements of language among the agents facing cognitive constraints and environmental pressures, but not explicitly rewarding cooperation or communication of a particular type \cite{bratman2010new}. These computational experiments are still far from being able to reproduce something a group of preschool children can do easily and habitually in the course of a play date \cite{Children_Rule_Making_Hardecker_2016}, \cite{piaget2015moral}. 

From an evolutionary perspective, the rules of the game developed by a group of intelligent agents should be stable enough to facilitate predictability in interactions, but flexible enough to facilitate adaptability and exploration. A fully orthodox, rigid set of rules inevitably fails in solving communal problems when environmental pressures change. A fully heterodox and easily malleable set of rules fails to facilitate the accumulation of knowledge and helpful heuristics.  A hierarchical set of behavioural constraints is helpful in achieving a balance between orthodoxy and heterodoxy:  rules of the game (the strictest set), norms (a softer set), and customs (the softest set). The trade-off between rule-making orthodoxy and heterodoxy is clearly task-, environment-, and situation-specific. Human societies exhibit various ways of solving it. 

Natural agents exhibit deceptive behaviours --  mostly instinctive. It is not self-evident that deceptive behaviours should be completely excluded from the design objectives of artificially intelligent agents. Elements of deception are already introduced by the fact that an intelligent agent does not have to disclose its own internal state (such as its intents) to other agents. In an environment populated by heterogeneous agents with partially conflicting objectives deception and concealment help agents achieve individual and group objectives  when they face the hostile actions of adversarial agents.

The social behaviour of intelligent agents necessarily includes a commitment to particular rules and strategies. It is not yet clear how the mechanism of commitment should work. Violation of rules can be rewarding and, indeed, for this reason, should be optimal if the "reward is enough" hypothesis is true for socially intelligent behaviours. "Reward is enough" cannot explain the existence of rules, norms and customs.

Rules require enforcement mechanisms. Norms and customs require support and maintenance. How these mechanisms of social order can emerge in a commune of social agents is not yet clear. There are conflicting hypotheses regarding how they evolve in human societies \cite{north2012violence}. 

Rule-making, development and enforcement of rule-enforcing mechanisms, within a single agent  (self-enforcing commitment) as well as between the agents, are likely key to understanding intelligent social behaviour. 

\section{Summary and where do we go from here}
A non-exhaustive list of interesting and open questions that follows from the discussion above: 
\begin{itemize}
    \item AI research will likely benefit from the demarcation of complex reflexive behaviours of reward-maximising artificial agents, on the one hand, and complex non-reflexive or intelligent behaviours which involve representations of reality and foresight, on the other hand.
    \item Singular reward maximisation has been able to model fairly complex reflexive and fixed action behaviours. Further progress in this area will likely answer the following questions: 
    \begin{itemize}
        \item What is the efficient method of incorporating multi-dimensional rewards and costs and soft and hard constraints specific to the agent's objective?
        \item How can the agent use various trade-off functions which specify the relative values of various components of costs and rewards?
        \item Are there efficient methods of incorporating human heuristics which impose or modify behaviours and help the agent handle situations in the areas of the state space sparsely populated by historical data?
        \item How do soft and hard constraints which force ethical norms and customs onto the agent's behaviours interact with multi-dimensional costs and rewards, trade-off functions, and heuristics?
    \end{itemize}
    \item Artificial agents tasked with complex objectives often operate in changing environments. The development of safe and robust artificial agents requires solving the contradiction between the data-centric character of the reward and cost and constraints heuristic optimization, on the one hand, and the variability of the agent's environment, on the other. An agent which is unable to recognize that the environment in which it operates right now is substantially different from the environment for which it has been trained -- is unsafe and is not robust. Quick adaptation imposes limitations on the agent's perceptual, cognitive and motor/action capacity.  What are the trade-offs between various aspects of adaptability such as the complexity of the agent's perception and cognitive machine, uncertainties and variability of the environment, and is there an efficient way to resolve them? Are there fundamental limits on the level of achievable optimality in data-driven learning in variable environments? If there are, perhaps a shift of focus from maximisation and optimization to achieving robustness and safety and a "good enough" but stable level of performance is justified. 
    \item The concept of commitment to a particular policy or sequence of actions is likely the key to solving the hierarchical behaviours which involve a top-level objective and multiple levels of sub-objectives. How can we introduce commitment in the structure of agents' rewards, costs, heuristics, trade-offs and constraints?
    \item Markov decision process and its modifications is often a gross oversimplification when an artificial agent is tasked with a complex, time-consuming objective. How can we allow the agent to store and recall salient truths about its own history such as crucial facts and situations and scenarios, not through its ability to optimise its expected rewards/costs in a local context, but because some truths, despite their rarity, are important \emph{instrumentally}? 
    \item The ability to choose behaviours and strategies in the presence of contingencies, and uncertainties, including deep uncertainties, is one of the characteristics of an intelligent agent. Trade-offs are always present in such circumstances. Resolving them is only partially possible with training and learning based on past data and by relying on human-imposed constraints and heuristics. The ability to foresee and to include in the agent's decision making novel, unseen, but possible scenarios is a requirement for a safe and robust artificial intelligent agent.
    \item Negotiation in the presence of commitments, making rules and creating behavioural and cognitive norms, as well as a mutual commitment to the norms and rules, their maintenance and adaptation are distinct characteristics of social intelligent agents. Can we systematically study rule-making or rule-improvement and rule-maintenance by social agents?
\end{itemize}

\section{Related work}

Sen \cite{sen1977rational} discusses utilitarian conceptions of intelligent beings and offers an extensive critique of the assumption of "self-interest in each act" as the definition of rational behaviour, and departures from it as irrational behaviour. 

McCloskey \cite{mccloskey_2016}, \cite{mccloskey_2017} offers an extensive critique of reward maximisation under constraints as the central mechanism shaping economic and social behaviours. 

A variant of the "reward is enough" hypothesis is discussed by Bostrom in the context of a "paperclip maximiser" in \cite{bostrom2003ethical}, \cite{bostrom2020ethical}, see also \cite{yudkowsky}.

A study \cite{abel2021expressivity} whether an appropriate Markov reward function can be constructed to capture some classes of desirable tasks has recently shown that this is not always the case.

Critique of the reward maximization principle and the link between \cite{SILVER2021103535} and behaviourism has been discussed in \cite{Roitblat_2021}.

\section{Conclusion}

The authors of "Reward is enough" hypothesise that the singular reward maximisation principle is the overarching truth behind a variety of phenomena associated with intelligence, natural and artificial. I have shown that reward maximisation is not enough to explain observations and is not helpful as an intellectual framework.

Reinforcement learning and reward maximisation have shown their power in a limited class of applications. Reward maximisation is insufficient as an intellectual framework to study intelligence and to create effective, safe, and robust artificial agents.

\section{Acknowledgments}

The author is grateful to his colleagues Hans Buehler, Maxence Hardy, Jonathan Kochems and Nicolas Marchesotti for thoughtful comments and stimulating discussions.

\bibliographystyle{elsarticle-num-names}

\bibliography{reward-ref.bib}

\end{document}